\documentclass[10pt, conference, compsocconf]{IEEEtran}
\usepackage{acronym}
\usepackage{xcolor}
\usepackage{multirow}

\usepackage{hyperref}

\ifCLASSINFOpdf
  \usepackage[pdftex]{graphicx}
\else
\fi
\usepackage[cmex10]{amsmath}
\usepackage{gensymb}

\usepackage{stfloats}

\usepackage{soul}

\newcommand{\AKadd}[1]{\textcolor{black}{ #1}}
\newcommand{\AKdel}[1]{}

\begin{document}
%
\title{LaserSAM: Zero-Shot Change Detection Using Visual Segmentation \\ of Spinning LiDAR}


\author{\IEEEauthorblockN{Alexander Krawciw}
\IEEEauthorblockA{Robotics Institute\\
University of Toronto\\
Toronto, Canada\\
alec.krawciw@mail.utoronto.ca}
\and
\IEEEauthorblockN{Sven Lilge}
\IEEEauthorblockA{Robotics Institute\\
University of Toronto\\
Toronto, Canada\\
sven.lilge@utoronto.ca}
\and
\IEEEauthorblockN{Timothy D. Barfoot}
\IEEEauthorblockA{Robotics Institute\\
University of Toronto\\
Toronto, Canada\\
tim.barfoot@utoronto.ca}
}


%


\maketitle

\begin{abstract}
This paper presents an approach for applying camera perception techniques to spinning LiDAR data. 
To improve the robustness of long-term change detection from a 3D LiDAR, range and intensity information are rendered into virtual perspectives using a pinhole camera model. 
Hue-saturation-value image encoding is used to colourize the images by range and near-IR intensity.
The LiDAR's active scene illumination makes it invariant to ambient brightness, which enables night-to-day change detection without additional processing. 
Using the range-colourized, perspective image allows existing foundation models to detect semantic regions.
Specifically, the Segment Anything Model detects semantically similar regions in both a previously acquired map and live view from a path-repeating robot.
By comparing the masks in both views, changes in the live scan are detected.
Results indicate that the Segment Anything Model accurately captures the shape of arbitrary changes introduced into scenes. 
\AKdel{The system achieves an object recall of 82.6\% and a precision of 47.0\%.}
\AKadd{The proposed method achieves a segmentation intersection over union of 73.3\% when evaluated in unstructured environments and 80.4\% when evaluated within the planning corridor.}
Changes can be detected reliably through day-to-night illumination variations. 
After pixel-level masks are generated, the one-to-one correspondence with 3D points means that the 2D masks can be used directly to recover the 3D location of the changes. 
\AKadd{The detected 3D changes are avoided in a closed loop by treating them as obstacles in a local motion planner. 
Experiments on an unmanned ground vehicle demonstrate the performance of the method. }
\end{abstract}


%
\IEEEpeerreviewmaketitle

\begin{figure}[t]
	\centering
	\includegraphics[width=0.9\linewidth]{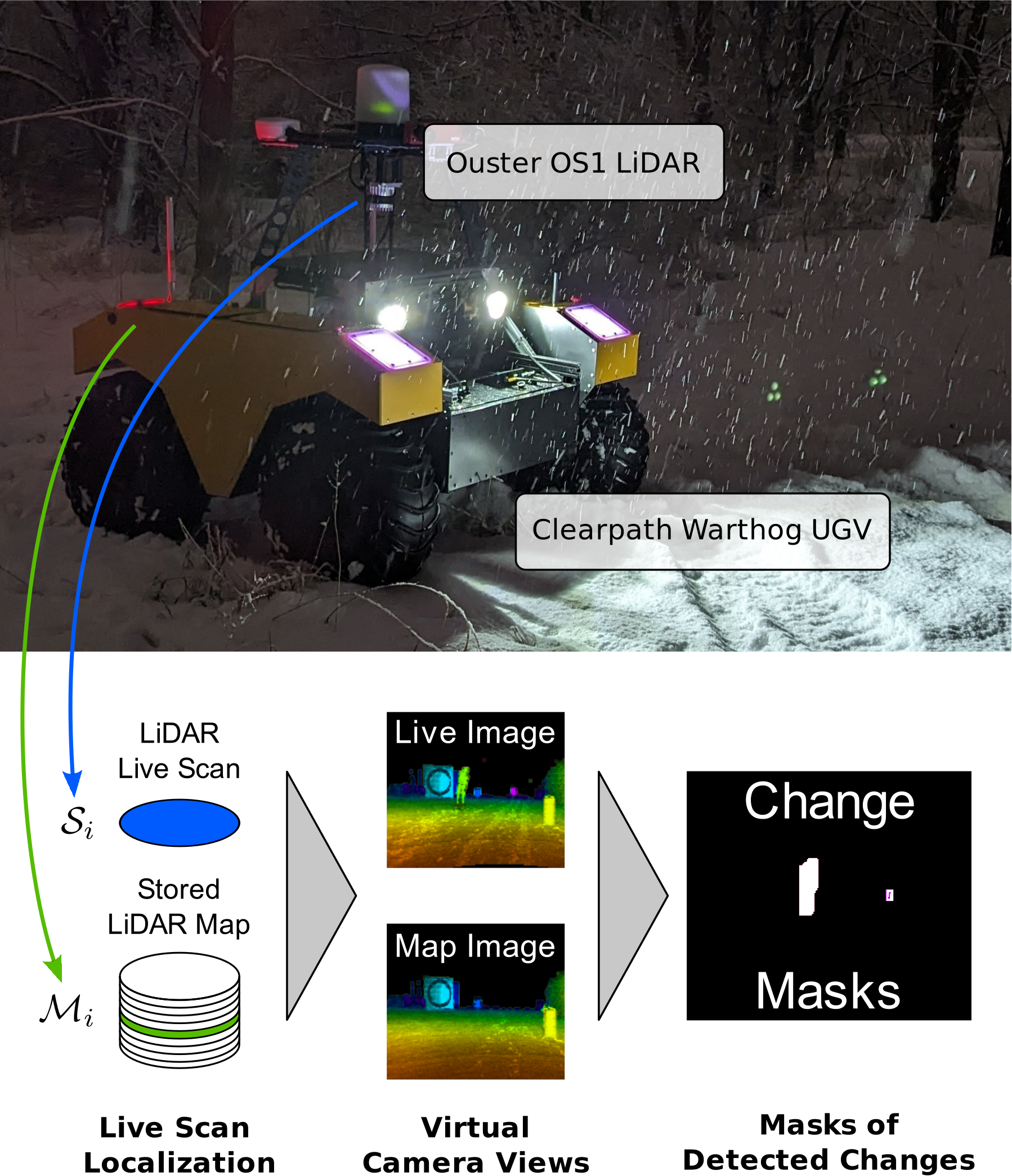}
	\caption{The Clearpath Warthog robot driving at night during snowfall, repeating a path that was previously taught during daytime. This paper proposes LaserSAM to detect and mask environmental changes between teach and repeat paths by creating virtual camera views from LiDAR data and applying \AKdel{modern }deep-learning-based segmentations.}
	\label{fig:warthigNight}
\end{figure}

\section{Introduction}
Robust perception remains a central challenge for mobile robots operating in unstructured environments. 
Unlike autonomous cars, there are no rules of the road and out-of-distribution classes of obstructions occur with a higher frequency \cite{ma_rethinking_2022}. 
Occlusion effects caused by vegetation degrade the quality of many detectors trained in structured environments \cite{qin_3d_2016}. 
These factors make it difficult for supervised perception pipelines to identify hazards reliably enough to maintain the aggressive performance of the vehicles. 
In addition to obstacle detection, off-road vehicles must perform terrain assessment to understand which local paths are feasible to drive \cite{papadakis_terrain_2013}. 
Visual Teach and Repeat (VT\&R) \cite{paul2010vtr} provides a practical alternative to terrain assessment. 
A human operator manually drives a robot along a network of connected paths.
While driving, the robot constructs local submaps, which are used later for localization. 
After the paths are taught, the robot can autonomously navigate between any two locations in the network.
The primary responsibility of terrain assessment is delegated to the human pilot, allowing aggressive maneuvers to be demonstrated to the robot. 
To allow safe repeats over long periods of time, we propose that it is sufficient to detect changes in the scene that intersect the planned local path of the vehicle \cite{deja_2016}, \cite{wu_thesis}. 
During the teach pass, the operator defines a safe corridor \cite{sehn_path} around the path.
When repeating, the robot navigates around new obstacles within the corridor's boundary. 
By focusing on detecting changes instead of specific types of hazards, any type of obstacle can be detected and avoided without predefining specific classes. 
For this reason, a robust change-detection module is desired to improve the system's capabilities.

Existing work on change detection has focused on images from monocular cameras \cite{cd_survey_05} and 3D LiDAR point clouds \cite{OKYAY2019}. 
Ding et al. \cite{ding2023adapting} use the FastSAM encoder as part of a visual change-detection approach for satellite images. 
Recently, change detection of point clouds for remote sensing \cite{de_gelis_siamese_2023} and mobile robotics \cite{krawciw} using deep learning has emerged as an area of study.
Conventional cameras have higher resolution than most spinning LiDAR sensors allowing them to detect shapes and textures more accurately. 
However, the lack of 3D information within the image makes it difficult to accurately transfer 2D segmentations into 3D for path planning \cite{qin_3d_2016}. 
Stereo vision systems can provide 3D information but must solve an additional data association problem.
While progress has been made to allow for direct comparison of images through lighting and seasonal changes \cite{chen_what_2023}, illumination variation complicates change detection for cameras.

Spinning $360\degree$ LiDAR scanners have emerged as a popular, complementary sensor to cameras \cite{wijayathunga_challenges_2023}.
LiDAR provides accurate 3D position and intensity measurements for each point. 
Most automotive LiDAR units operate at a wavelength of 1550 nm because the atmosphere absorbs most of the sun's energy at that wavelength \cite{ouster_howlidar}.
Ouster LiDARs operate at 840 nm allowing them to capture ambient sunlight \cite{ouster_howlidar}. 
The combination of ambient and projected infrared light leads to more natural illumination effects that blur the line between passive camera and active LiDAR. 

This work blends the strengths of both modalities by rendering perspective camera images from LiDAR scans.
This approach allows computer vision algorithms to be applied to multi-view change detection. 
Specifically, the Segment Anything Model (SAM) \cite{sam_fb} \AKdel{and the Fast Segment Anything Model (FastSAM) are} \AKadd{is} used to detect semantic regions in the rendered images.
\AKadd{Its zero-shot generalization capabilities allow for the segmentation of any object, including unseen ones in new domains.}
Multi-modal algorithms rely on extrinsic calibration between LiDAR and cameras to define the depth of some pixels in a camera image.
In this approach, every pixel in the generated image corresponds exactly to a 3D point.
3D points from the map and the live scan are rendered into a common virtual camera frame for analysis. 
Finally, the active illumination and signal processing of the 840 nm IR means that change detection works across illumination conditions with no additional processing required. 
\autoref{fig:warthigNight} demonstrates the types of conditions that challenge camera-based detection but do not impact the proposed LiDAR pipeline.

In summary, we propose the following contribution: a change detection pipeline that combines the sensor benefits of LiDAR with a pre-trained foundation model for image segmentation.


\section{Related Work}

\subsection{Change Detection}

Detecting changes in images, point clouds, or other rich sensors is often a key task in scene understanding \cite{cd_survey_05}.
One of the central challenges is to capture the domain-specific semantics accurately.
Differentiating inconsequential changes such as illumination and sensor orientation allows for meaningful downstream processing. 
Ultimately, these definitions must be determined at the problem level but the binary definitions of \textit{changed} and \textit{unchanged} are common in the literature \cite{cd_survey_05}. 
Most methods operate at the point/pixel level.
Aggregation into objects may not be necessary or occur as a downstream processing task.

\subsubsection{3D LiDAR Change Detection}
Change detection in point clouds is most commonly applied in remote sensing \cite{OKYAY2019} to detect large-scale changes such as new buildings \cite{de_gelis_siamese_2023}.
Classical methods are usually based on the geometry of the scene. The simplest point-cloud difference evaluates the distance of every point to its nearest neighbour \cite{girardeau2005change} and thresholds distant points as \textit{changed}. 
These thresholds can be place-dependent \cite{deja_2016} to adapt to the local geometry.
Alternative approaches use ray-tracing \cite{underwood_explicit_2013} or normal distances \cite{wu_thesis} to classify points based on fixed or variable thresholds.
Deep learning is successful in the change detection of 3D point clouds as well.
de Gelis et al. use parallel encoders to train both a supervised \cite{de_gelis_siamese_2023} and unsupervised \cite{de_gelis_unsup_2023} neural network that classifies points as added, removed, or the same. 
These works operate on point cloud maps that are constructed from many different sensor viewpoints, such as the SHREC 2023 dataset \cite{gao_shrec_2023}. 
This makes them less applicable to mobile robots, which are heavily affected by occlusions.

\subsubsection{2D Camera Change Detection}

Camera change detection can be classified in many ways, but the most relevant detail to this work is whether or not the method assumes a quasi-static background.
Quasi-static methods tend to construct a representation of the scene background that is static \cite{subsense_charles} or adaptive \cite{cd_survey_05}. 
These methods perform well in video surveillance.
Viewpoint variations from mobile robots make them less applicable to the data analyzed here. 
Fewer methods consider the temporal information from video data and attempt to track changes from a moving camera \cite{toyama_99}. 
This is difficult in the monocular case because the 3D position of the objects is unknown and warping an image into a new frame requires assumptions about the scene or camera motion. 

\subsection{Zero-Shot Semantic Segmentation}
In 2023, Meta Research released the Segment Anything Model (SAM) \cite{sam_fb}, which aims to perform semantic segmentation on any scene. 
The model is trained on over eleven million images with more than one billion masks.
This vast dataset has enabled the capability to segment regions in a class-agnostic manner. 
In contrast to common datasets for autonomous driving such as CityScapes \cite{Cordts2016Cityscapes} or SemanticKITTI \cite{behley2019iccv}, this allows the model to generate masks \AKdel{on tasks and cameras on which it was not trained.}
\AKadd{for images that were captured in new environments with different cameras.}
Additionally, the model decoder accepts geometric prompts such as points of interest or bounding boxes that refine the masks that are reported. 
While this performance is impressive, its run time is slow for robotics applications. 

\AKdel{
In response to this runtime constraint, FastSAM was developed based on the YoloV8  architecture.
It is trained on the same dataset as Meta's SAM but the simpler convolutional neural network architecture is much faster than the transformer model used originally.
However, the key tradeoff for the speed boost is prompt quality. 
FastSAM provides a list of masks and confidences but does not consider any additional prompts such as position or bounding boxes.
The impacts of this trade-off will be discussed later in the paper. }

\subsection{LiDAR as a Camera}
The most common LiDAR sensors used in \AKdel{modern } autonomous vehicle development are 360$\degree$ units from Ouster (Velodyne) \cite{ouster_128}, and Hesai \cite{hesai}.
These sensors are characterized by higher resolution along the horizontal spinning axis than in the vertical field of view (FOV). 
The current state-of-the-art sensors have 128 vertical beams and operate at 10 to 20 Hz \cite{ouster_128}. 
However, low-frame-rate LiDAR sensors have existed with much higher resolutions for more than two decades.
McManus et al. \cite{mcmanus_towards_2013} showed that lighting-invariant visual odometry could be performed using SURF features on intensity images taken from an Autonosys 2D scanning LiDAR.
This sensor produces images with a $30\degree \text{V} \times 90\degree$H FOV, with resolution $480 \times 360$ pixels at 2 Hz.
The frame rate is much slower, but the vertical resolution is still significantly higher (12 pixels per degree) than the OS-1 LiDAR (2.85 pixels per degree) used in this paper.
This approach was extended to account for the motion distortion caused by the moving sensor \cite{anderson2013ransac} and used to control an offroad vehicle in closed loop \cite{mcmanus2013lighting}, \cite{Barfoot2013IntoDV}.

\begin{figure*}
    \centering
    \includegraphics[width=0.98\textwidth]{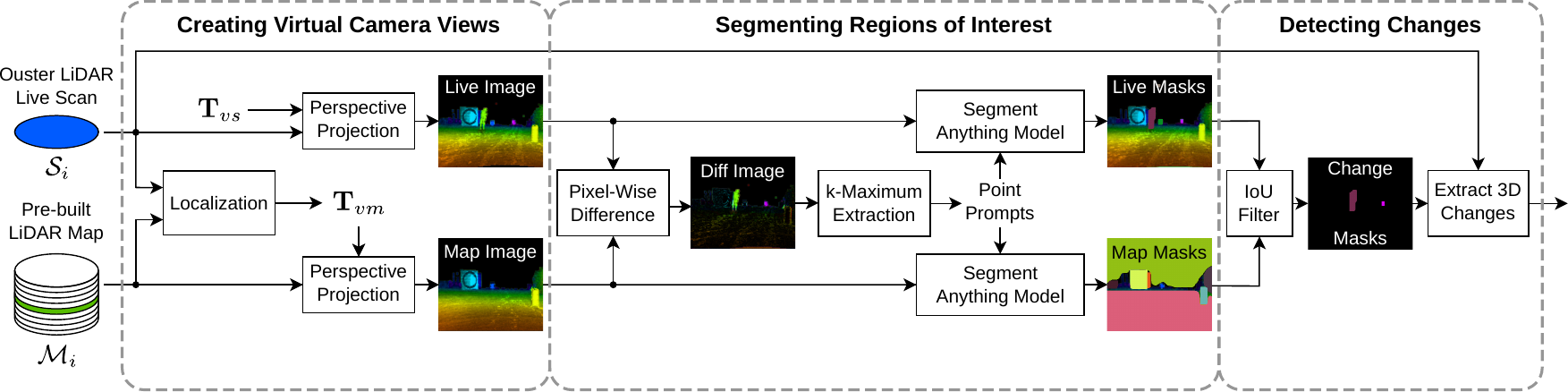}
    \caption{Data processing pipeline of LaserSAM. The pipeline runs for each new frame obtained from the Ouster LiDAR.}
    \label{fig:pipeline}
\end{figure*}

\section{Method}

This paper ties together past progress in change detection and LiDAR-based image processing with state-of-the-art foundation models.
By leveraging the strengths of traditional approaches, and the high-quality segmentation generated by SAM \cite{sam_fb}, our system can detect previously unseen obstacles, improving the autonomy of a mobile robot.

The proposed change-detection pipeline has three primary stages: creating virtual camera views, segmenting regions of interest, and detecting changes. 
These stages, as well as the internal steps, are illustrated in \autoref{fig:pipeline}.
The problem inputs are two point clouds: the local submap used in localization $\mathcal{M}_i$ and the live LiDAR scan $\mathcal{S}_i$.

\subsection{Creating Virtual Camera Views}
\begin{figure}
    \centering
    \includegraphics[width=0.49\textwidth]{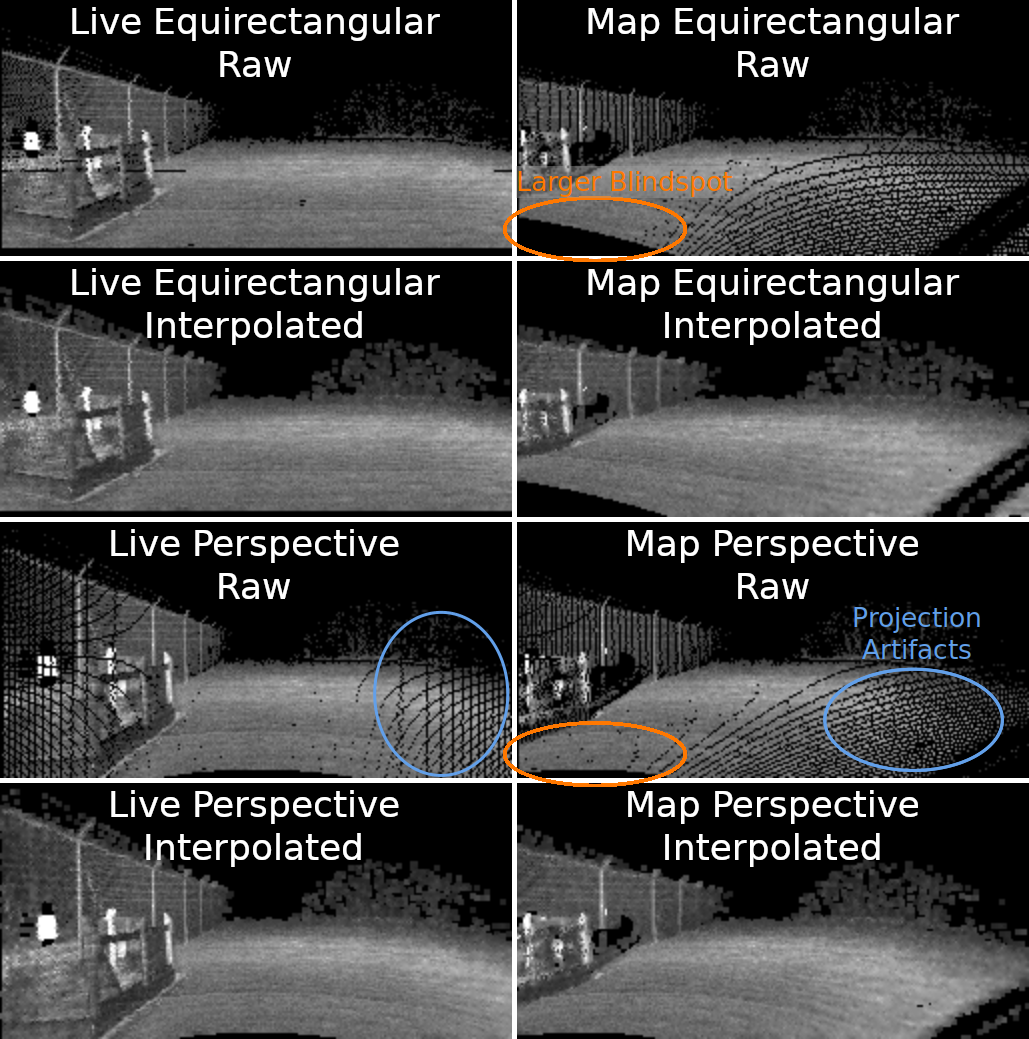}
    \caption{The left column contains equirectangular and perspective images aligned with the LiDAR. 
    The right column shows the two projections with a two-meter lateral offset.
    The perspective view has a smaller blind spot around the base of the robot.}
    \label{fig:processingPipeline}
\end{figure}

The core idea of this pipeline is to project the two aligned point clouds into images, the format expected by SAM.
The natural data format of a spinning LiDAR is equirectangular where each pixel corresponds to a constant angular offset. 
The natural data format of cameras is the perspective projection.
Sample images created using each projection are shown in \autoref{fig:processingPipeline}.
A larger blindspot around the base of the robot is created in the equirectangular view than in the perspective view when deviating from the path (the map view).
In the following, we will use the perspective projection because the image segmentation neural networks were trained on data in this form.  
We define a virtual camera pose in the vehicle's frame, $\textbf{T}_{cv} \in SE(3)$, which is used to render the images. 
The desired virtual camera pose is arbitrary, but aligning its position with the origin of the LiDAR sensor minimizes the amount of interpolation required in the final image.  
The first step is to align the point clouds in the camera frame.
When a new live scan is received, an existing ICP-based localization module \cite{burnett_are_2022} is used to extract the relative pose of the vehicle in the map ($\textbf{T}_{vm} \in SE(3)$).
The extrinsic transformation between the LiDAR sensor and the vehicle, $\textbf{T}_{vs} \in SE(3)$, is known from calibration.

The monocular pinhole camera model \cite{Barfoot_2024} is used to render the virtual camera image in the camera frame. By convention, the $z$-axis extends forward from the camera.
The pixel positions are 
\begin{equation}
    \begin{bmatrix}
        u_j \\
        v_j
    \end{bmatrix} = \mathbf{g}(\mathbf{p}_j) = 
    \begin{bmatrix}
        f_u & 0 & c_u \\
        0 & f_v & c_v
    \end{bmatrix} \\
    \frac{1}{z} \begin{bmatrix}
        x \\
        y \\
        z
    \end{bmatrix}.
\end{equation}
The image is the union of the intensity values for each pixel in 
the map point cloud after it is transformed into the camera's frame:
\begin{equation}
    I_\text{map} = \left\{ \begin{bmatrix}
        u_j \\
        v_j \end{bmatrix} = \mathbf{g}(\textbf{T}_{cv}\textbf{T}_{v m_i}\mathbf{p}_{m_i}^j) \big| \mathbf{p}_{m_i}^j \in \mathcal{M}_i
    \right\}.
\end{equation}
Similarly, the live scan is
\begin{equation}
    I_\text{live} = \left\{ \begin{bmatrix}
        u_j \\
        v_j \end{bmatrix} = \mathbf{g}(\textbf{T}_{cv}\textbf{T}_{vs}\mathbf{p}_{s_i}^j) \big| \mathbf{p}_{s_i}^j \in \mathcal{S}_i
    \right\}.
\end{equation}

The image width ($W$) and height ($H$) are set as $256 \times 128$ pixels based on the resolution of the OS-1 sensor. 
Horizontal and vertical fields of view are set to be $\text{fov}_V \times \text{fov}_H = 90\degree \times 45\degree$.
An ideal, centred camera is used, leading to
\begin{subequations}
\begin{align}
    c_u = W/2, \\
    c_v = H/2.
\end{align}
\end{subequations}
The focal length is defined based on the desired image size and field of view as 
\begin{subequations}
    \begin{align}
        f_u = \frac{W}{2 \tan (\text{fov}_H / 2)}, \\
        f_v = \frac{H}{2 \tan (\text{fov}_V / 2)}.        
    \end{align}
\end{subequations}
Points beyond the field of view are ignored, and if multiple points are captured by the same pixel, the closest point to the camera is retained. 

When a perspective projection is used, gaps in the image are created due to the distortion. 
These gaps are highlighted in the third row of \autoref{fig:processingPipeline}.
Interpolating these regions is critical for the segmentation to work properly. 
A $3 \times 3$ kernel is used to interpolate the missing pixels, but only non-zero pixels are considered.
The distortion is exacerbated by any offset of the virtual camera's origin from the LiDAR. 
Conceptually, the virtual camera could be colocated with the LiDAR's position in either the teach or repeat.
However, we choose to locate it in the repeat frame because the map point clouds are an accumulation of points from sequential scans. 
The hue-saturation-value colour model is used to colour the images by range. 
The $z$ coordinate in the camera frame is used for hue, mapping over a maximum range of 30 m.
The saturation is the constant 255 for all pixels and the value is the LiDAR intensity of the pixel.
The first column of \autoref{fig:qualResults} shows two sample images rendered from the robot in a common frame.

\subsection{Segmenting Regions of Interest}
After rendering both images in a common frame, segmentation occurs.
\AKdel{
We compare the performance of SAM and FastSAM for this step.
SAM can perform segmentation based on point prompts of interest. 
To generate point prompts, the norm of the pixel-wise difference of the two images is evaluated. 
The top $k$ local maxima of the resulting 2D surface are used as prompts to the model.
The model is prompted twice, once using the live image and once using the map image. 
This generates two lists, each containing $k$ masks from the live and map images, respectively.
FastSAM cannot accept prompts in the same way as SAM. 
Instead, the model is evaluated independently for each image.
This means that there are two lists of masks at each timestep.
The two lists of masks are independent meaning their lengths often differ and a correspondence problem presents itself to associate masks in teach and repeat.}
\AKadd{The pre-trained model \texttt{sam\_vit\_b} \cite{sam_fb} 
is used without fine-tuning for all experiments.}
SAM requires a geometric prompt on the image to define the anchor point(s) of the mask. 
To remain as general as possible, we use a single point-prompt for each semantic mask. 
Prompt selection is a critical aspect of the method.
Each prompt adds run time to the model so we are constrained to select a few prompts, rather than a uniform grid that covers the whole image. 
To select prompts, \AKadd{two methods were compared. The first uses} the norm of the pixel difference between the live image and the map image.
The top-$k$ local maxima are selected as prompts. Additionally, a minimum distance constraint is imposed to ensure greater image coverage.
\AKadd{The second method uses the $k$-largest centroids of connected components of the 3D nearest neighbours changes projected into the camera view.}

\subsection{Detecting Changes}
After independently segmenting the image pairs, the detected masks are compared.
To decide if two masks capture the same object, the intersection-over-union (IoU) score is calculated between them.
The bitwise AND ($\wedge$) and OR ($\vee$) operations are used to calculate the IoU of two binary masks ($b1, b2$),
\begin{equation}
    \text{IoU} = \frac{\sum_i \sum_j b1[i][j] \bigwedge b2[i][j]}{\sum_i \sum_j b1[i][j] \bigvee b2[i][j]}.
\end{equation}
An IoU score of one indicates a perfect match and zero means no overlap between the individual masks.
Masks with a maximum IoU of less than 0.5 are considered to be \textit{changed} between the live scan and the map.
\AKadd{A secondary 3D check is used to ensure that the point clouds intersect in 3D as well as in their projection.}
The remaining masks are considered \textit{unchanged}.
\AKdel{For SAM, this is a direct comparison of the matched lists.
For FastSAM, a brute-force comparison between every mask in the two scans is required. 
There are typically fewer than ten masks so this is computationally acceptable. }
All points classified as \textit{changed} in the live scan are then back-projected into 3D. 
The 3D bounding box, centroid, or raw points can be estimated and provided to a local planner.
A corridor-constrained sampling-based planner \cite{sehn_path} inflates the \textit{changed} points to account for the robot's footprint and avoids them.

\renewcommand{\arraystretch}{1.25}

\begin{table*}[b]
\centering
\caption{\AKadd{Point-Wise Comparison of LaserSAM to Change Detection Baseline Algorithms.}}
\label{tab:quantResults}
\begin{tabular}{|c|c|c|c|c|c|c|c|} 
\cline{2-7}
\multicolumn{1}{l|}{}   & \multicolumn{3}{c|}{Full Field of View}                & \multicolumn{3}{c|}{Corridor Filtered}                 & \multicolumn{1}{l}{}    \\ 
\hline
\textbf{Method}           & \textbf{IoU}    & \textbf{Precision} & \textbf{Recall} & \textbf{IoU}    & \textbf{Precision} & \textbf{Recall} & \textbf{Run Time (ms)}  \\ 
\hline
            Pixel Difference Baseline       & 13.7\%          & 15.3\%             & 55.9\%          & 21.5\%          & 28.0 \%            & 48.0\%         & 5.7 $\pm$ 1.2           \\ 
\hline
LaserSAM     with      Pixel Difference Prompts & 28.2\%          & 39.7\%             & 49.4 \%          & 47.1\%          & 68.0\%            & 60.5\%          & 281.3 $\pm$ 27.9        \\ 
\hline
             3D Difference Baseline           & 56.5\%          & \textbf{95.5\% }   & 58.0\%          & 58.6\%         & \textbf{97.0\%}   & 59.7\%         & 80.6 $\pm$ 13.2         \\ 
\hline
LaserSAM            with 3D Difference Prompts    & \textbf{73.3\%} & 86.5\%             & \textbf{82.8\%} & \textbf{80.4\%} & 94.4\%             & \textbf{84.5\%} & 356.2 $\pm$ 30.9        \\
\hline
\end{tabular}
\end{table*}

\section{Experiments}

\begin{figure*}
    \centering
    \includegraphics[width=0.8\textwidth]{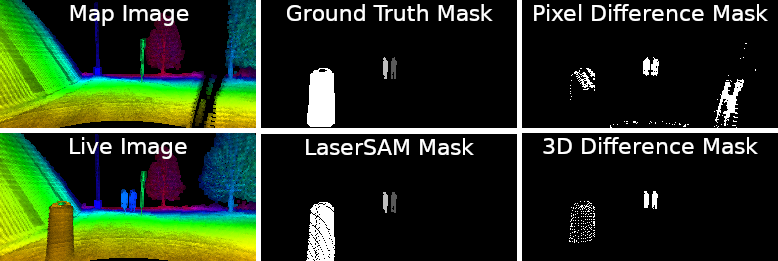}
    \caption{Change-detection masks for a sample frame. The left column shows the input images. There are three changed objects, two pedestrians and a static cone, marked in the ground truth. The segmentation results for each algorithm are shown in their respective panel.}
    \vspace{-3mm}
    \label{fig:qualResults}
\end{figure*}

Experiments were performed at the University of Toronto's Institute for Aerospace Studies using a Clearpath Warthog UGV \cite{warthog} equipped with an Ouster OS-1 128 beam LiDAR \cite{ouster_128}. 
Data were recorded during night and day as well as during snowstorm conditions.
Static obstacles placed on and around the original path and pedestrians walking near the robot were introduced as changes in the dataset.
A mixture of natural and fabricated items was used to minimize the impact of the chosen materials' reflectivity on the experiments.
Additionally, duplicate items were placed near the path as part of the static scene \AKadd{during mapping}.
This demonstrates that the algorithm is performing change detection, not obstacle classification.
Two different sequences were recorded with changes.
The first sequence is 262 m long on relatively flat terrain.
The second is 230 m through a wooded area. 
Changes to be detected include cones, mannequins, pedestrians and road signs.
\AKdel{The forest sequence is notable because the teach pass occurred during daylight and the repeat pass was recorded in the dark, about six hours later. }

\AKadd{The dataset has 12,012 live scans to use for testing. 
One hundred frames were randomly selected for manual annotation to provide ground truth masks. }
The ground-truth masks were instance segmented to allow for per-object metrics to be evaluated in addition to per-point ones. 
As discussed in the related work, the precise semantic meaning of a change is problem-dependent.
In this dataset, only new or moved obstacles that would impact the robot's ability to drive were marked as changed. 
Small scene changes such as footprints in the snow, or tracks from previous drives were marked as \textit{unchanged}.

\subsection{Results}


The pipeline was evaluated on the annotated subset of the frames. 
\AKdel{Specifically, SAM and FastSAM were compared to the baseline pixel-wise difference algorithm, which was used to create the point prompts for SAM.}
\AKadd{Two baselines were selected for comparison: a camera-style pixel difference method and a 3D-geometric detector with a Gaussian roughness model \cite{wu_thesis}. 
The pixel-wise IoU between the predicted and ground-truth segmentation masks is used for comparison.}
\AKdel{Two metrics are used for comparison: the pixel-wise IoU between the predicted and ground-truth segmentation masks and the F-score of objects being detected.}
\AKadd{To quantify the effectiveness of each method for robot navigation, the detection performance is restricted to be within the allowable planning corridor.
This planning-oriented metric is useful because errors in change detection that will not block the robot are not as critical to safe operation.}
\AKdel{The ground truth is instance-segmented, which allows for the evaluation to consider individual changes in the same frame, and whether or not they are detected. 
An overlap of 50\% IoU is the threshold for a true positive detection for object-based metrics.
After counting the number of true positives (TP), false positives (FP), and false negatives (FN) detections, the precision, recall, and F-score can be evaluated []. }
The achieved results are summarized in \autoref{tab:quantResults}.
All computational timing was performed on a laptop NVIDIA RTX A4500 GPU and an Intel i7-12800H CPU. 
\AKadd{LaserSAM was compared to both baselines, using the five most likely regions of change from each baseline as the input prompts.
When prompted using pixel differences, LaserSAM significantly reduces the number of false positives, as demonstrated by the increase in precision.
When prompted using 3D distances, there are fewer false positives but the mask-refining capabilities more accurately capture the whole object.
This improvement is clear from the recall increase from 59.7\% to 84.5\% after corridor filtering.}


\autoref{fig:qualResults} shows the segmentation result for a sample frame with two moving pedestrians and a static cone. 
\AKadd{LaserSAM generates the most accurate segmentation mask, provided that it has been prompted effectively.
The gaps in the traffic cone correspond to projection artifacts. 
While interpolation is used to generate the mask, only those pixels that correspond to a 3D measurement are retained as changes.
A secondary benefit of LaserSAM is the instance groupings of points. 
In \autoref{fig:qualResults}, each instance is coloured according to the ground truth instance that it matches most closely. 
The baseline methods provide only binary labels for each point in the scan. }
\AKdel{All three methods correctly find the three objects but FastSAM marks extra objects, and SAM has the most accurate segmentation boundary. 
In our experiments, SAM outperforms both other methods by a significant margin. }

\begin{figure}
    \centering
    \includegraphics[width=0.48\textwidth]{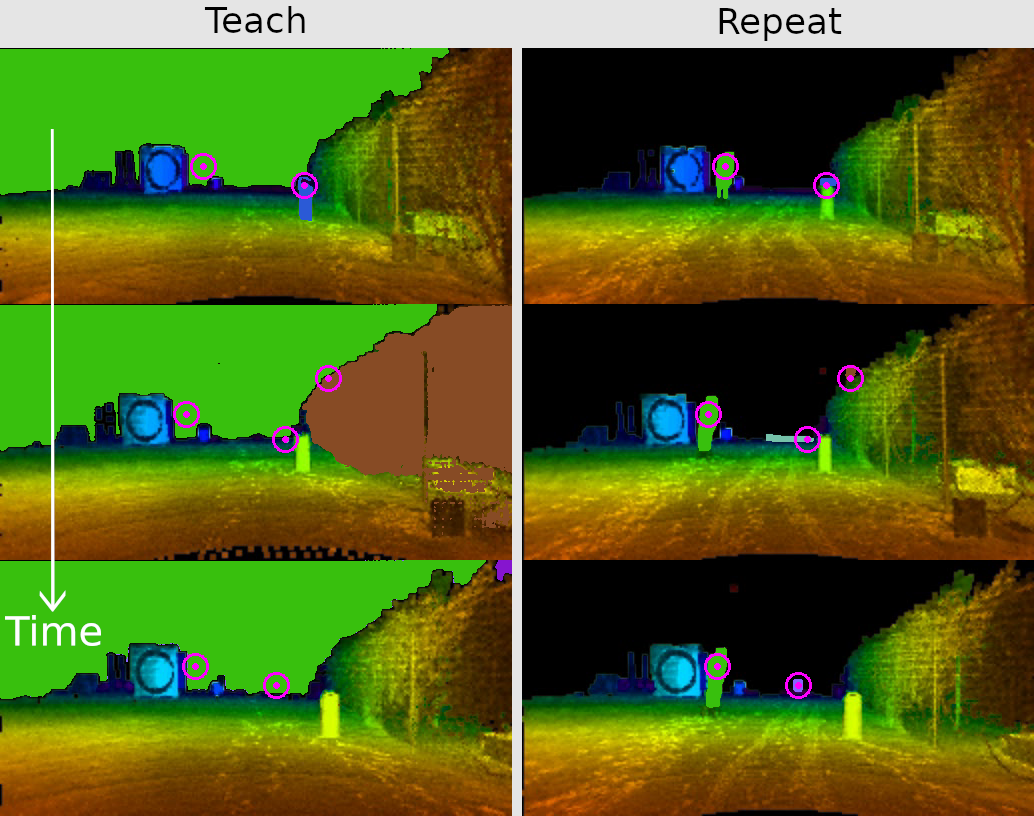}
    \caption{A temporal sequence of semantic regions generated by SAM \cite{sam_fb}. The shared prompt point is highlighted in the pink bulls-eye. The mask colours match between the teach and repeat. }
    \label{fig:samSeq}
    \vspace{-10px}
\end{figure}

\AKdel{
In its current state, the system has a high recall, but low precision; there are too many false positives. 
Qualitative analysis provides some additional insight into the results. 
SAM and FastSAM tend to over-segment in failure cases.
The most commonly observed failure mode during our experiments is a localization error, causing a large difference on the static object region boundary.
In Fig 1, live and map images are not prompted inside the same object so SAM segments large, non-overlapping regions.
This has an outsized impact on the IoU because many false-positive pixels are added to the IoU calculation.
On the object level, the results are more encouraging. In successful frames, such as Fig X., the changes are captured accurately and there are no isolated bright pixels common in the baseline or 3D point clouds approaches.
We anticipate that with improved localization, the number of large, static false-positive detections will fall rapidly. }

\AKdel{ provides a sample set of three matched teach and repeat frames extracted using FastSAM.
Critically, the changed objects are classified correctly within the scene.
However, some masks flicker and are not retained as desired. 
Sometimes the ground is not masked as a region in the teach pass leading to an error where the entire ground may be considered a change in the repeat.
While post-processing filters this out based on size, ground errors near the robot have the largest impact on the path planner downstream. }

\autoref{fig:samSeq} shows an example sequence of masks generated by SAM. 
When changes are detected, the map image often has a large mask corresponding to the ground or sky. 
The masks from SAM tend to be consistent over time which is advantageous for robot motion planning. 
Prompts often exist near the boundaries, which is undesirable.
Once the two masks are generated, the IoU threshold filter suppresses false positives.

\subsection{Closed-Loop Experiments on Warthog UGV}
LaserSAM was packaged into C++ and integrated into the Visual Teach and Repeat 3 framework\footnote{\href{http://github.com/utiasASRL/vtr3}{github.com/utiasASRL/vtr3}}.
The run time of LaserSAM with 3D prompting is $356 \pm 30.9$ ms, which is slower than real-time for the 10 Hz OS-1 LiDAR. 
A performance compromise is achieved by running lidar odometry at 10 Hz but only running change detection as frequently as possible.
The 2 - 3 Hz update rate for the local costmap is fast enough to avoid static and slow-moving obstacles. 
Much of the computation occurs on the GPU allowing odometry on the CPU to proceed in parallel. 
The LiDAR sensor has a near-range blindspot so the planner maintains previously detected changes in a queue, which adds temporal permanence that is not provided by LaserSAM.

\begin{figure}
    \centering
    \includegraphics[width=0.49\textwidth]{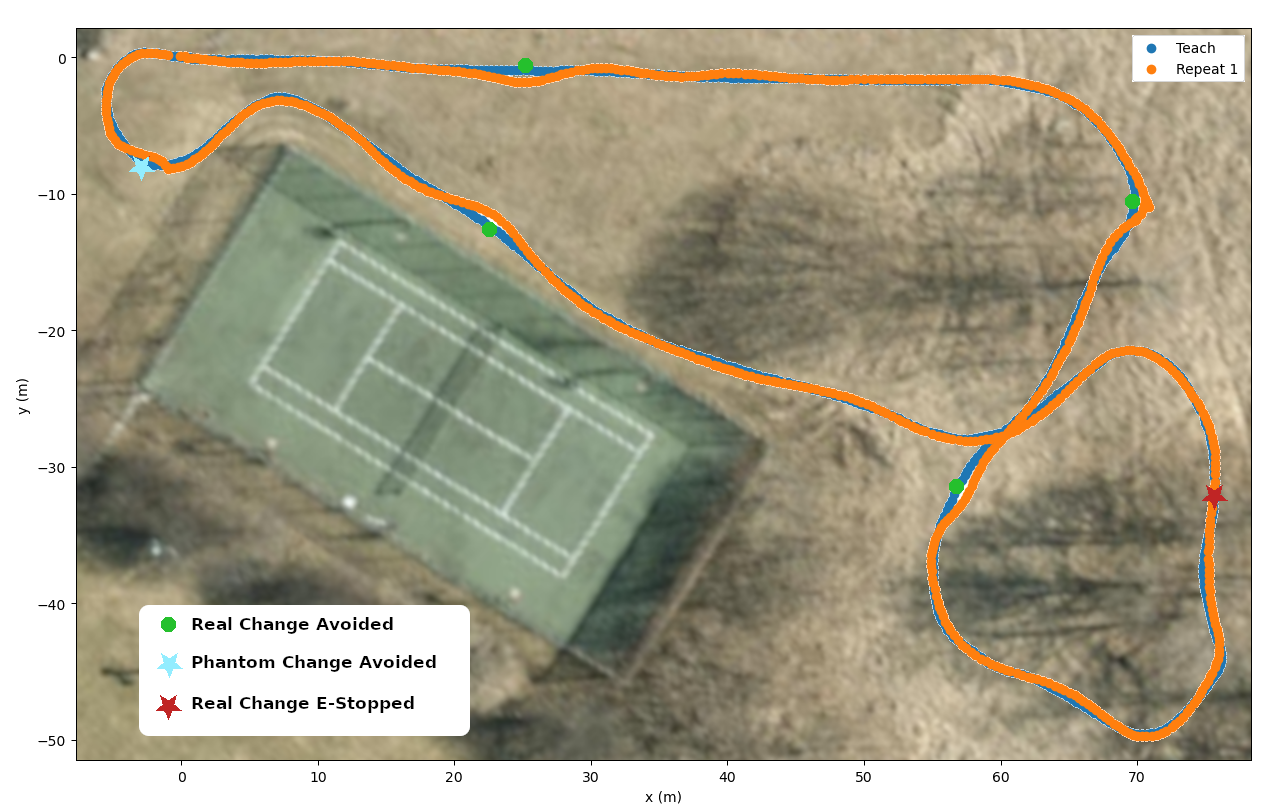}
    \caption{The path of the robot in orange as it detects and avoids changes in real time using LaserSAM. The blue line represents the obstacle-free teach path.}
    \label{fig:closedLoop}
\end{figure}

\begin{figure}[b]
    \centering
    \includegraphics[width=0.45\textwidth]{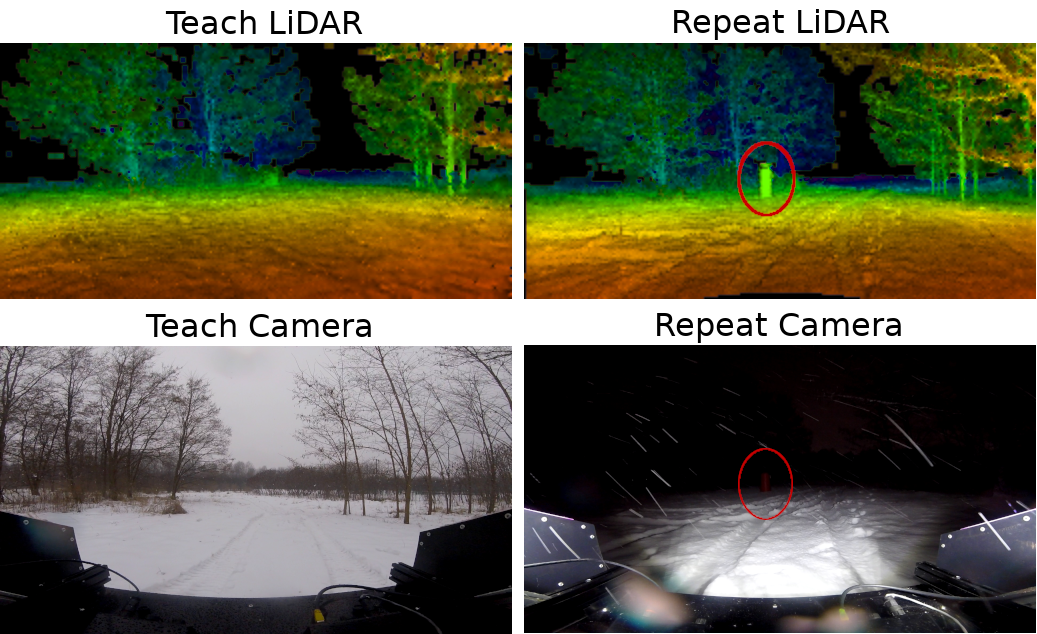}
    \caption{The rendered LiDAR images are not affected by ambient lighting conditions. The teach was performed during sunlight and the repeat was performed in the dark, and during heavier snow.}
    \label{fig:nightDay}
\end{figure}

\begin{figure*}
    \centering
    \includegraphics[width=0.7\textwidth]{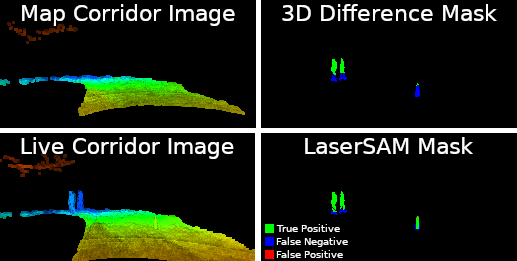}
    \caption{The change detection quality in the planned corridor between the 3D Difference Baseline and LaserSAM. LaserSAM uses semantic information to capture points closer to the object boundary with the map. This leads to more green pixels on all three objects near the ground.}
    \label{fig:IoUView}
\end{figure*}


\AKadd{
In the sample trajectory shown in \autoref{fig:closedLoop}, the robot's obstacle-free path is drawn in blue and is mostly covered by the orange repeat.
The robot veers around changes that are detected along the path.
Four changes are avoided correctly, and the robot returns to the obstacle-free path afterwards. 
One small cone was not detected and an emergency stop from the operator was required before continuing. 
Finally, in one location there was a false positive detection that caused an unnecessary evasive maneuver.}

\AKadd{
In another experiment, the path was repeated six hours after the mapping was performed after night had fallen.
LaserSAM's performance is unaffected by the ambient illumination of the scene.
\autoref{fig:nightDay} shows the two virtual images in the afternoon and night.
The colouring of the images is qualitatively indistinguishable.
This stands in contrast to the conventional camera to which the cone is not visible.
An attempt to use SAM for change detection on the RGB camera would fail under these conditions.}

\AKadd{
The two primary benefits of using segmentation over the baselines are a better definition of changes near the map and the suppression of noisy detections caused by mapping artifacts. 
\autoref{fig:IoUView} shows a pedestrian that is partially segmented based on 3D distances, but using the SAM mask instead, the detection reaches the ground interface. 
Changes smaller than the 3D baseline's distance threshold can be detected more accurately with LaserSAM because the intensity data defines their boundaries better than distances alone. }

\subsection{Limitations}
\AKdel{Existing foundation models are trained on colour RGB images, not greyscale nor HSV range-colourized images. 
The extracted features are likely less expressive than the network's capability. }

\begin{figure}
    \centering
    \includegraphics[width=0.45\textwidth]{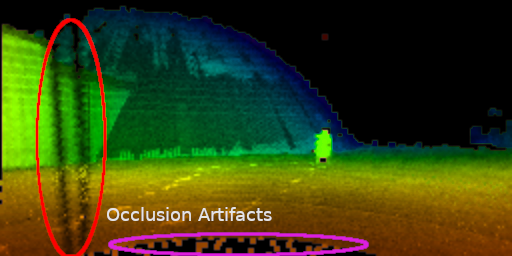}
    \caption{Range-coloured image highlighting the effect of the occlusions and submaps. In the red ellipse, a shadow in the scan is visible from the support strut of the robot that blocked the LiDAR in the teach. In the magenta ellipse, points are visible in the local blindspot of the sensor because the submap contains a union of points from multiple scans. }
    \vspace{-10px}
    \label{fig:occlusions}
\end{figure}
Similar to works that process point clouds directly, occluded regions remain a challenge.
\autoref{fig:occlusions} is rendered from a different pose than the LiDAR meaning that occluded parts of the scene are visible from the new view. 
This causes a shadow effect from the structure of the robot's sensor supports.
The further from the LiDAR's origin that the virtual camera is rendered, the more gaps there are in the visual frame.
\AKadd{In closed-loop testing, this creates an undesirable feedback loop because when the robot deviates from the path to avoid obstacles detection accuracy can decrease.}
The submap has fewer occlusion-based gaps because it is constructed from sequential frames. 
Points within the magenta region of \autoref{fig:occlusions} were appended to the local submap during the teaching process, filling in some of the holes caused by displacing the virtual camera.
If storage was unlimited, retaining every point from nearby teach frames would allow for denser maps and minimize occlusion effects.
In teach and repeat, the robot's lateral path displacements are typically less than two meters, which minimizes these effects in practice.
\AKdel{A final challenge involves the accuracy of the localization algorithm. 
If the relative transformation between the two point clouds is inaccurate, the rendered images are misaligned and the intersection of the masks check will fail.
In the perspective view, small angular misalignments have a large impact on the image location of distant changes. Small translational errors, particularly along the $z$-axis of the camera, impact the position of nearby changes. }

\subsection{Future Work}
\AKdel{
Currently, the method has only been evaluated on one type of LiDAR sensor.
However, it would be interesting to understand how much the intensity quality impacts overall performance.
A potential ablation study could add increasing levels of noise to the intensity data, eventually resulting in uniform white noise coloured only by range. }
To improve temporal consistency and add tracking, the centroid of previous obstacles will be projected into the camera at the next timestep and used as a prompt. Changes will likely remain visible in the scene, so prompting on their estimated location should improve temporal consistency.
Inconsistent masks could be filtered out as spurious.
Lastly, this work only considers positive changes: items added in the live scan. 
It should be possible to detect items that existed in the teach pass on the path that were removed.
In a manual mapping process, it is unlikely that a robot will be driven over items that can disappear so this is left for future investigation.

\section{Conclusion}
This paper demonstrates that developments in the intensity sensitivity of LiDAR scanners enable algorithms that leverage the benefits of multi-modal detection but use only a single LiDAR sensor. 
While the resolution of commodity LiDAR-rendered images lags behind conventional cameras, it has reached sufficient thresholds to apply vision algorithms to field robots. 
\AKadd{
The zero-shot generalization capabilities of the Segment Anything Model allow for visual change detection to be applied to simulated camera images created from LiDAR data. 
On the test set, using LaserSAM with 3D-based prompting achieves an IoU of 73.3\%. When considering only obstacles within the allowable planning corridor performance is higher with an IoU of 80.4\%.
With LaserSAM running on a Clearpath Warthog UGV, static changes can be detected and avoided reliably. 
The operational frequency of 2.8 Hz is fast enough for driving up to 1 m/s. 
By suppressing spurious false-positive detections the local planner can find better routes. 
LaserSAM naturally works through illumination changes from night to day and day to night. 
Continued efficiencies in prompt generation and optimization for inference should help decrease the run time and allow LaserSAM to detect dynamic changes. }
\AKdel{Although the 3.1 Hz Segment Anything Model is slower than the 73 Hz FastSAM, the ability to prompt masks based on the live pixel-level difference in intensity and the previously viewed obstacles makes it more effective than the FastSAM frame-by-frame approach. 
The classic trade-off between speed and effectiveness leaves both foundation models as viable methods for future evaluation.
We look forward to building on these results; implementing the neural network live on the robot and running it in closed-loop with our path planner.}




\newcommand{\BIBdecl}{\setlength{\itemsep}{0.25 em}}
\bibliographystyle{IEEEtran}
\bibliography{IEEEabrv,refs}
%

\end{document}